\documentclass[runningheads]{llncs}
\usepackage[T1]{fontenc}
\usepackage{graphicx}
\usepackage{tikz}
\usepackage{array}
\usepackage{amsmath}
\usepackage{tabularx}
\usepackage{multirow}
\usepackage{url}

\newcolumntype{Y}{>{\centering\arraybackslash}X}

\begin{document}
\title{
Analysis of 
Respiratory Sinus Arrhythmia
\\with Neural Networks}

\author{
Julian Szyma\'nski\inst{1} \orcidID{0000-0001-5029-6768}, \\
Patryk Orkisz\inst{1}\orcidID{0009-0008-2272-6423}, \\
Higinio Mora\inst{2}\orcidID{0000-0002-8591-0710}\\
}

\institute{
    Gdansk University of Technology, \\ Narutowicza 11/12, 80-233 Gdansk, Poland \\
    \email{\{julian.szymanski, patryk.orkisz\}@pg.edu.pl}
     \and
     University of Alicante, \\ 03690 San Vicente del Raspeig, Alicante; Spain;
    \\
     \email{hmora@dtic.ua.es}
}

\maketitle              
\begin{abstract}

The paper introduces a neural network-based approach for analyzing ECG signals to estimate respiratory rate by leveraging the phenomenon of Respiratory Sinus Arrhythmia (RSA). Our method employs a deep learning model trained to predict respiratory waveforms directly from ECG input data. To achieve this, we developed and evaluated three different neural network architectures capable of automatically extracting relevant features from ECG signals without the need for manual preprocessing. The proposed approach offers a robust and scalable solution for non-invasive respiratory monitoring, with potential applications in healthcare and wearable technology.


\end{abstract}

\section{Introduction}

The variability of vital parameters, such as heart and respiratory rates, is an indicator of overall health. This variability reflects the body's ability to proper respond to external stimuli and adaptation to surrounding conditions. Electrocardiography (ECG) is one of the fundamental diagnostic tools for assessing cardiovascular health and is used to identify various conditions, ranging from acute and chronic myocardial injury to cardiac arrhythmias, structural heart disease, and inflammatory processes. 
 

ECG is not only used to evaluate physical health but also to assess mental well-being. This is because the heart is influenced by the sympathetic and parasympathetic branches of the autonomic nervous system (ANS). ANS regulates the function of various organs, including the endocrine and respiratory systems, in response to states of alertness or relaxation. By leveraging the complementary nature of these two ANS branches, biomarkers of both physical and mental health can be determined. Through ECG signal analysis using QRS complex detectors or R-wave detection, time-domain metrics of heart rate variability (HRV) can be derived. 
Additionally, frequency-domain HRV metrics can be obtained through spectral analysis of the ECG signal using Fourier transform methods.
\cite{shaffer2017overview}.

Another non-obvious physiological indicator that can be extracted from an ECG is the respiratory rate. By analyzing heart rate variability (HRV), specifically the time differences between consecutive heartbeats (known as RR intervals), ECG-Derived Respiration (EDR) can be obtained. EDR refers to the extraction of respiratory information from ECG and serves as a noninvasive method for monitoring respiratory activity in cases where direct respiratory signals are not recorded. This is possible due to the interactions between the cardiovascular and respiratory systems, a phenomenon known as Respiratory Sinus Arrhythmia (RSA)
\cite{yasuma2004respiratory}.
RSA is a natural physiological variation in heart rate linked to the phases of respiration. In a synchronized manner, the intervals between consecutive RR beats fluctuate with the breathing cycle. This interaction between the two branches of the ANS is mediated by the vagus nerve, which influences heart activity. During inhalation, vagal activity decreases, causing the heart rate to accelerate. Conversely, during exhalation, vagal activity increases, leading to a slower heart rate.
 
Traditional ECG signal processing algorithms have limitations, such as task-specific design, manual feature extraction, and sensitivity to noise and patient variability. Their multi-stage processes also make them complex and less adaptable. In contrast, neural network-based algorithms outperform traditional methods, as demonstrated in numerous studies, by adapting, learning continuously, and handling multiple tasks. They autonomously identify relevant features, making them more robust to noise and patient variations.
In this research, we propose a neural network algorithm for ECG signal analysis, enabling the estimation of HRV and RSA. This approach outperforms traditional methods, particularly in noisy conditions, where its effectiveness becomes more apparent. Additionally, it allows for easy model personalization through the integration of individual patient data.

The rest of this paper is structured as follows. Section 2 provides an overview of ECG signal processing methods used for prediction and regression tasks, with a particular focus on deep learning algorithms. Section 3 presents our 
approach to processing ECG and respiratory signals, 
describing in detail the neural network architecture employed for R-wave detection and nasal cannula signal regression. Section 4 discusses the performance of the model in detecting R-wave peaks and examines the results of experiments on respiratory rate prediction. It also describes the training and testing process of the respiratory regression network and compares the proposed algorithm's effectiveness with a classical method implemented in the NeuroKit library. Finally, 
section 5 summarizes the obtained results and outlines directions for future research.
 
\section{ECG Signal Analysis Methods} 
In ECG signal processing and analysis tasks, three groups 
of algorithms can be distinguished \cite{SOTA_ECG}—Traditional Algorithms, Machine Learning Algorithms and Deep Learning Algorithms.
Algorithms from these groups are used in tasks such as anomaly detection, classification, or signal delineation—segmentation of cardiac cycles into regions \cite{serhani2020ecg}.

The first group includes rule-based methods or traditional signal processing approaches, which involve feature extraction and classification stages for these tasks Typically, the first step is filtering using a filter-based approach (low pass, band pass, adaptive, etc.). Then, time-domain and morphological features are extracted, usually based on signal windowing algorithms or transforms. The final step is classification, which uses thresholding-based algorithms. 

The next group consists of classical machine learning algorithms.
\cite{SOTA_ECG} describes the applications of methods such as Support Vector Machines (SVM), Decision Trees, Random Forest, Naïve Bayes, and K-Nearest Neighbors (KNN) in ECG analysis.

The last group of algorithms employs neural networks. In the literature, a variety of approaches to ECG analysis using neural networks can be found \cite{petmezas2022state}, using different architectures, starting with those based on dense layers, moving through recurrent and convolutional layers, and ending with transformer architectures \cite{liu2021deep}. In the wide range of applications of neural networks for ECG processing, we have outlined below the approaches related to HRV and EDR analysis.

In the study \cite{Qin_z}, a dense network was used to assess human stress levels based on HRV analysis by selecting reliable HRV features from the time and frequency domains and classifying stress states. In contrast, the study \cite{ST_RES_NET} implemented a convolutional network with a U-NET architecture to detect entire QRS complexes and R-wave peaks. An enhancement to the classic U-NET encoder-decoder architecture was proposed by incorporating spatiotemporal feature extraction blocks.
A less complex architecture was presented in \cite{robust_r_peak_detection_lstm}, where two recurrent layers were used to classify ECG signal samples to identify R-wave peaks. The study \cite{HRV-Transformer} introduced the HRV-Transformer model for contactless heart rate measurement using a camera. To enable real-time heart rate detection, part of the Transformer decoder architecture was removed.

In \cite{QRS_T_localization}, the authors conducted a comprehensive review of QRS complex and ST-T segment localization.
They categorized the methods into traditional and Deep Learning approaches and proposed an Ensemble Training-based algorithm. They trained two U-NET-based models: a convolutional and a recurrent network.

The literature rarely addresses the respiratory waveform or pattern regression task, especially using end-to-end deep learning methods. In most studies, researchers train models using datasets collected through photoplethysmography (PPG) or other biosensors, such as capnography (which measures $CO_2$ concentration in inhaled and exhaled air).
In the paper \cite{DL_rr_from_biosignals}, the authors used three different datasets: Capnobase, MIMIC-II, and sEMG—to train and test various neural network architectures designed for the task of predicting respiratory flow using one-step-ahead prediction. Subsequently, the peaks from the predicted impedance signals are extracted by identifying the maximum from a set of extrema points within small signal windows. 
In \cite{DL_rr_from_physiological_signals}, the authors conducted a literature review on the application of machine learning methods for determining the RR metric, confirming that this is not a widely explored research area. They proposed an LSTM model and compared its performance with an ARIMA model in the task of respiratory time series forecasting using the BIDMC dataset.
The study \cite{rr_neural_models} introduced a respiratory rate prediction algorithm that leverages both convolutional and recurrent layers. Additionally, it presented other architectures such as RRWaveNet, RespWatch, CycleGAN, and RespNet, which utilize convolutional and dense layers or their combination with recurrent layers. These architectures are characterized by a high number of parameters. In most cases, existing neural models are employed for RR metric prediction rather than for respiratory waveform regression.

\section{Neural Model for RSA Analysis} \label{section/method}
Code - \url{https://github.com/patrykcommander/rsa_analysis}
 
To train a neural network model for regressing ECG signals to respiratory waveforms from a nasal cannula, a dedicated dataset was used, collected using the Aidmed One device by Aidlab \cite{Aidlab}. This device recorded single-channel ECG waveforms along with the respiratory signal from the nasal cannula. Preprocessing is applied to both types of signals, with the ECG preprocessing discussed first.

 
 
Initially, a function that applies convolution to the one-dimensional ECG signal was used to smooth the signal. The length of the row vector, determined by the parameter $ kernel\_length = 5$, contained convolution kernel weights, each with a value of $1/kernel\_length$.

The second step was made only for waveforms from our dataset.
The data was annotated using the \textit{ecg\_peaks} function from the \textit{NeuroKit} package.
In contrast, the MIT-BIH and Fantasia datasets already contain pre-existing annotations. 
The WFDB package \cite{WFDB} was used for processing these public datasets.

In the next step, the ECG annotations were transformed.
Originally, the annotations were a vector of integers representing the sample indices at which R-peaks are present. A new vector, matching the shape of the corresponding ECG, was created, where each sample represents the likelihood of an R-peak occurring at that position.

The ECG waveforms and the transformed annotations were then divided into windows of equal duration, set to 10 seconds. The windowing function included an additional parameter, $overlap\_factor$, which defines the degree of overlap between adjacent windows.
 
Next, the signal is filtered to remove noise with a wavelet transform. 
Then the preprocessed signal window is normalized to the range $\langle -1, 1 \rangle$ using min-max normalization. Subsequently, downsampling has been performed for datasets with a sampling frequency greater than 100 Hz.

\begin{figure}
	\centering
	\scalebox{0.7}{

\begin{tikzpicture}
\definecolor{emb_color}{RGB}{252,224,225}
\definecolor{multi_head_attention_color}{RGB}{252,226,187}
\definecolor{add_norm_color}{RGB}{242,243,193}
\definecolor{ff_color}{RGB}{194,232,247}
\definecolor{softmax_color}{RGB}{203,231,207}
\definecolor{linear_color}{RGB}{220,223,240}
\definecolor{gray_bbox_color}{RGB}{243,243,244}
\definecolor{conv_layer}{RGB}{240,100,100}
\draw[fill=gray_bbox_color, line width=0.046875cm, rounded corners=0.300000cm] (-0.975000, 6.455000) -- (2.725000, 6.455000) -- (2.725000, 1.305000) -- (-0.975000, 1.305000) -- cycle;
\draw[line width=0.046875cm, fill=emb_color, rounded corners=0.100000cm] (0.000000, 0.000000) -- (2.500000, 0.000000) -- (2.500000, -0.900000) -- (0.000000, -0.900000) -- cycle;
\node[text width=2.500000cm, align=center] at (1.250000,-0.450000) {CNN \vspace{-0.05cm} \linebreak network};
\draw[line width=0.046875cm, fill=add_norm_color, rounded corners=0.100000cm] (0.000000, 3.680000) -- (2.500000, 3.680000) -- (2.500000, 3.180000) -- (0.000000, 3.180000) -- cycle;
\node[text width=2.500000cm, align=center] at (1.250000,3.430000) {Add \& Norm};
\draw[line width=0.046875cm, fill=multi_head_attention_color, rounded corners=0.100000cm] (0.000000, 3.030000) -- (2.500000, 3.030000) -- (2.500000, 2.130000) -- (0.000000, 2.130000) -- cycle;
\node[text width=2.500000cm, align=center] at (1.250000,2.580000) {Multi-Head \vspace{-0.05cm} \linebreak Attention};
\draw[line width=0.046875cm, fill=add_norm_color, rounded corners=0.100000cm] (0.000000, 6.230000) -- (2.500000, 6.230000) -- (2.500000, 5.730000) -- (0.000000, 5.730000) -- cycle;
\node[text width=2.500000cm, align=center] at (1.250000,5.980000) {Add \& Norm};
\draw[line width=0.046875cm, fill=ff_color, rounded corners=0.100000cm] (0.000000, 5.580000) -- (2.500000, 5.580000) -- (2.500000, 4.680000) -- (0.000000, 4.680000) -- cycle;
\node[text width=2.500000cm, align=center] at (1.250000,5.130000) {Feed \vspace{-0.05cm} \linebreak Forward};
\draw[line width=0.046875cm] (1.250000, 5.580000) -- (1.250000, 5.730000);
\draw[line width=0.046875cm] (1.250000, 0.600000) circle (0.200000);
\draw[line width=0.046875cm] (1.410000, 0.600000) -- (1.090000, 0.600000);
\draw[line width=0.046875cm] (1.250000, 0.760000) -- (1.250000, 0.440000);
\draw[line width=0.046875cm] (0.350000, 0.600000) circle (0.400000);
\draw[line width=0.046875cm] (-0.030000, 0.600000) -- (-0.014490, 0.629156) -- (0.001020, 0.657833) -- (0.016531, 0.685561) -- (0.032041, 0.711884) -- (0.047551, 0.736369) -- (0.063061, 0.758616) -- (0.078571, 0.778258) -- (0.094082, 0.794973) -- (0.109592, 0.808486) -- (0.125102, 0.818576) -- (0.140612, 0.825077) -- (0.156122, 0.827883) -- (0.171633, 0.826946) -- (0.187143, 0.822284) -- (0.202653, 0.813971) -- (0.218163, 0.802145) -- (0.233673, 0.786999) -- (0.249184, 0.768783) -- (0.264694, 0.747796) -- (0.280204, 0.724382) -- (0.295714, 0.698925) -- (0.311224, 0.671845) -- (0.326735, 0.643584) -- (0.342245, 0.614608) -- (0.357755, 0.585392) -- (0.373265, 0.556416) -- (0.388776, 0.528155) -- (0.404286, 0.501075) -- (0.419796, 0.475618) -- (0.435306, 0.452204) -- (0.450816, 0.431217) -- (0.466327, 0.413001) -- (0.481837, 0.397855) -- (0.497347, 0.386029) -- (0.512857, 0.377716) -- (0.528367, 0.373054) -- (0.543878, 0.372117) -- (0.559388, 0.374923) -- (0.574898, 0.381424) -- (0.590408, 0.391514) -- (0.605918, 0.405027) -- (0.621429, 0.421742) -- (0.636939, 0.441384) -- (0.652449, 0.463631) -- (0.667959, 0.488116) -- (0.683469, 0.514439) -- (0.698980, 0.542167) -- (0.714490, 0.570844) -- (0.730000, 0.600000);
\draw[line width=0.046875cm, -latex] (1.250000, 3.680000) -- (1.250000, 4.680000);
\draw[line width=0.046875cm, -latex] (1.250000, 0.000000) -- (1.250000, 0.400000);
\draw[line width=0.046875cm, -latex] (1.250000, 0.800000) -- (1.250000, 2.130000);
\draw[line width=0.046875cm] (0.750000, 0.600000) -- (1.050000, 0.600000);
\draw[-latex, line width=0.046875cm, rounded corners=0.200000cm] (1.250000, 4.080000) -- (-0.750000, 4.080000) -- (-0.750000, 5.980000) -- (0.000000, 5.980000);
\draw[-latex, line width=0.046875cm, rounded corners=0.200000cm] (1.250000, 1.530000) -- (-0.750000, 1.530000) -- (-0.750000, 3.430000) -- (0.000000, 3.430000);
\draw[-latex, line width=0.046875cm, rounded corners=0.200000cm] (1.250000, 1.730000) -- (0.312500, 1.730000) -- (0.312500, 2.130000);
\draw[-latex, line width=0.046875cm, rounded corners=0.200000cm] (1.250000, 1.730000) -- (2.187500, 1.730000) -- (2.187500, 2.130000);
\draw[line width=0.046875cm, -latex] (1.250000, -1.500000) -- (1.250000, -0.900000);
\node[text width=2.500000cm, anchor=north, align=center] at (1.250000,-1.500000) {$X$};
\node[anchor=east] at (-1.175000,3.880000) {$N\times$};
\node[text width=2.000000cm, anchor=east] at (0.250000,0.600000) {Positional \vspace{-0.05cm} \linebreak Encoding};

\node[line width=0.046875cm, fill=add_norm_color, rounded corners=0.100000cm, minimum width=2.5cm, minimum height=0.5cm, draw=black] at (1.250000,7.3000) (dense) {Linear layer};

\draw[-latex, line width=0.046875cm] (dense) -- (1.250000, 8.1);

\draw[-latex, line width=0.046875cm] (1.250000,6.2000) -- (dense);

\node[text width=2.500000cm, anchor=north, align=center] at (1.250000,8.55) {$Y$};

\node[rectangle, rounded corners=0.2cm, fill=emb_color, draw=black, line width=0.046875cm, minimum width=4cm, minimum height=6.25cm] at (6.5, 3.25) (embedding) {};

\node[rectangle, line width=0.046875cm, fill=conv_layer, rounded corners=0.100000cm, draw=black, minimum width=2.5cm, minimum height=1cm] at (6.5, 1.25) (conv_1) {Conv 3x1, 8};

\node[rectangle, line width=0.046875cm, fill=conv_layer, rounded corners=0.100000cm, draw=black, minimum width=2.5cm, minimum height=1cm] at (6.5, 3.25) (conv_2) {Conv 3x1, 16};

\node[rectangle, line width=0.046875cm, fill=conv_layer, rounded corners=0.100000cm, draw=black, minimum width=2.5cm, minimum height=1cm] at (6.5, 5.25) (conv_3) {Conv 3x1, 32};

\draw[-latex, line width=0.046875cm] (conv_1) -- (conv_2);
\draw[-latex, line width=0.046875cm] (conv_2) -- (conv_3);

\draw[-latex, line width=0.046875cm] (conv_3) -- (6.5, 7);
\draw[-latex, line width=0.046875cm] (6.5, -0.5) -- (conv_1);

\node[text width=2.500000cm, anchor=north, align=center] at (6.5, -0.65) {$X$};

\node[anchor=north, align=center] at (6.5, 7.55) {Embedding};
\end{tikzpicture}

}
	\caption{Our model, based on a Transformer network, used for ECG signal classification.}
	\label{img/ecg_transformer}
\end{figure}
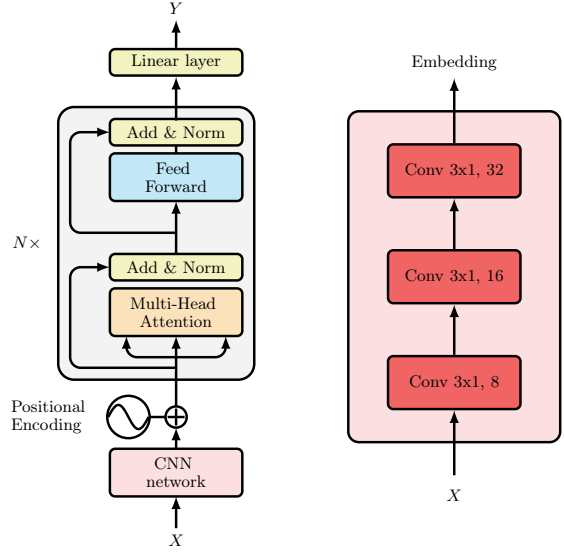

The processed data was used for training, validation, and testing of three neural network architectures dedicated to R-peak detection. The first is based on recurrent layers \cite{robust_r_peak_detection_lstm}, the second on a U-Net architecture \cite{ST_RES_NET}, and the last is designed using a Transformer encoder. The Transformer-based model is presented in Fig. \ref{img/ecg_transformer}.

Three CNN layers are used to increase the number of channels in the ECG input data. Then, sine and cosine positional encoding is applied. The Transformer encoder block is created using the following parameters: $input\_dim = 32$, $num\_layers = 2$, $num\_heads = 4$, $dim\_feedforward = 32$, $dropout = 0.2$. Finally, a linear layer with a sigmoid activation function is applied.

Using metrics aligned with AAMI standards \cite{AAMI} and performance criteria, we decide to use in our further research model based on recurrent layers \cite{robust_r_peak_detection_lstm}.

\begin{figure}
	\centering
	\includegraphics[scale=0.5]{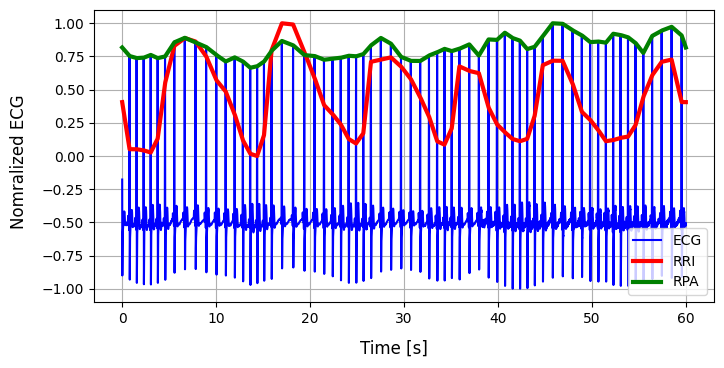}
	\caption{ECG waveforms, along with interpolated RRI and RPA}
	\label{img/ecg_new_channels}
\end{figure}

The last step prepares the data for the model used for respiratory rhythm reconstruction, using the previously trained detector. It involves extending the input signal with two additional information: RRI (time intervals between R-peaks) and RPA (amplitudes of R-peaks). To adjust the dimension of the additional vectors to match the length of the ECG, linear interpolation was applied. Next, the RRI channel is normalized using min-max normalization to the range <0, 1>. The prepared data, along with the preprocessed respiratory waveforms, were used to design a neural network for respiratory regression. Fig. \ref{img/ecg_new_channels} shows the result of introducing the additional channels. This serves as the input data to the neural network.

Following the ECG preprocessing, the respiratory waveform preprocessing is described next.
The data were processed using the NeuroKit and Physio libraries \cite{Physio}\cite{goldberger2000physiobank}.

Initially, annotations were obtained semi-automatically, then manually verified and corrected, serving as reference data for testing.

The next step involves synchronizing the ECG and respiratory signal time series. 
If timestamp differences occur at the sequence's start or end, the signals are trimmed accordingly.

Then, the entire sequence has been smoothed using a convolution function with a parameter \textit{kernel\_length} = 100.

In the next step, the transformation of the respiratory signal annotation vectors and the partition of the resulting time series into smaller windows were performed. This was done in a similar way to the ECG waveforms. As a result, windows of the respiratory signal, ECG, respiratory peak annotations, and respiratory cycle annotations were obtained.

Each of the resulting respiratory windows was processed using a Butterworth band-pass filter, with cutoff frequencies of 0.05 Hz and 3 Hz. The final operation is the transformation using the \textit{log\_softmax} function, which ensures the appropriate value range for the loss function criterion used in the model.

\begin{figure}
	\centering
	\includegraphics[scale=0.45]{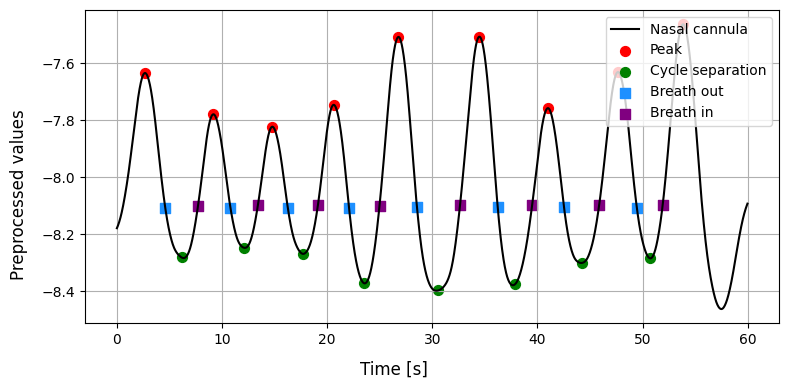}
	\caption{Respiratory signal annotations}
	\label{img/adnotacje_resp}
\end{figure}

Fig. \ref{img/adnotacje_resp} shows an example of a window from the cannula signal after initial preprocessing.
In the first plot, the signal peaks, obtained using the NeuroKit library, are marked in red. Then, using the Physio tool, the exhalation and inhalation phases are marked in light blue and purple, respectively, while the dark green points represent the average values of the exhalation and inhalation indices.
 
The prepared in such a way dataset was filtered to remove signal windows in which the maximum respiratory frequency was below 
9 breaths per minute. This step is associated with limiting the RSA phenomenon \cite{denver2007methodological}. Typically, RSA is strongly 
related to the high-frequency band (HF), which corresponds to a breathing rate of approximately 9-24 breaths per minute \cite{Nagarajan_RSA}. However, when the breathing rate exceeds a certain threshold, the RSA amplitude starts to decrease. 
At breathing frequencies above around 0.12 Hz (approximately 7.2 breaths per minute), RSA starts vanishing. With very fast breaths, such as during hyperventilation, the heart stops dynamically responds to the respiratory cycle, and RSA may become inconsistent \cite{Niizeki_K}. This indicates that there is a point at which the heart stops responding to respiration.

\begin{figure}
	\centering
	\scalebox{0.5}{

\begin{tikzpicture}
    \definecolor{linear_color}{RGB}{194,232,247} 
    \definecolor{lstm_color}{RGB}{250,214,165}   
    \definecolor{norm_color}{RGB}{210,247,194}   
    \definecolor{interpolate_color}{RGB}{230,230,230} 

    \tikzstyle{layer} = [rectangle, minimum width=5cm, minimum height=1.2cm, font=\Large, thick, rounded corners=2mm, draw]
    \tikzstyle{arrow} = [-latex, thick, shorten >= 0.1cm, shorten <= 0.1cm, line width=1mm]
    \tikzstyle{external} = [circle, minimum size=1cm, font=\Large, thick, draw]
    \tikzstyle{description} = [draw=none, font=\Large, align=left, minimum width=2cm, anchor=west]

    \node[external] at (0,14) (input) {$X$};

    \node[layer, fill=lstm_color, align=center] at (0,12) (lstm_1) { Bidirectional LSTM};
    \node[layer, fill=norm_color, align=center] at (0,9.8) (layer_norm_1) { Layer normalization};
    \node[layer, fill=linear_color, align=center] at (0,7.6) (dense_1) { Linear layer};
    \node[layer, fill=interpolate_color, align=center] at (0,5.4) (interpolate) { Interpolation layer};
    \node[layer, fill=lstm_color, align=center] at (0,3.2) (lstm_2) { Bidirectional LSTM};
    \node[layer, fill=linear_color, align=center] at (0,1) (dense_2) { Linear layer};

    \node[external] at (0,-1) (output) {$Y$};

    \draw[arrow] (input) -- (lstm_1);
    \draw[arrow] (lstm_1) -- (layer_norm_1);
    \draw[arrow] (layer_norm_1) -- (dense_1);
    \draw[arrow] (dense_1) -- (interpolate);
    \draw[arrow] (interpolate) -- (lstm_2);
    \draw[arrow] (lstm_2) -- (dense_2);
    \draw[arrow] (dense_2) -- (output);


    \node[description] at (4, 14) {\textbf{Parameters:}};

    \node[description] at (4, 12) {input\_size $=$ 3 \\ hidden\_size $=$ 64 \\ dropout $=$ 0,6};

    \node[description] at (4, 9.8) {normalized\_shape $=$ 128};

    \node[description] at (4, 7.6) {in\_features $=$ 128 \\ out\_freatures $=$ 64};

    \node[description] at (4, 5.4) {interpolate\_ratio $=$ 0,2 \\ mode $=$ 'linear'};

    \node[description] at (4, 3.2) {input\_size $=$ 64 \\ hidden\_size $=$ 64};

    \node[description] at (4, 1) {in\_features $=$ 128 \\ out\_freatures $=$ 1};

\end{tikzpicture}
}
	\caption{Our neural model for respiratory rhythm regression based on ECG signal}
	\label{img/ecg_to_resp}
\end{figure}
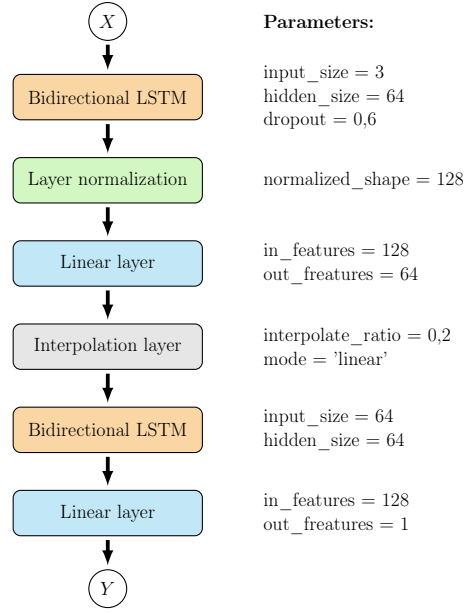

Our model for respiratory regression from electrocardiographic data is presented in Fig. \ref{img/ecg_to_resp}. 
It consists of two bidirectional LSTM layers, two linear layers, a single normalization layer, and a single interpolation layer, which was introduced to account for the nonidentical sampling frequencies of the ECG and respiratory signals. The model was trained using our preprocessed dataset, with a nominal sampling frequency. Initially, a weighted sum of the absolute error criterion ($L1$) and the normalized cross-correlation ($NCC$) coefficient was used as the loss function. However, the best training results in the experiments were achieved using the Kullback-Leibler ($D_{KL}$) divergence criterion in combination with the absolute error
(Formula \ref{loss_function}) 
, where $\mathcal{L}$ - loss function, $y$ - target, $y_{pred}$ - regression, $\alpha = 0.9$, $P$ - probability distribution of the ground truth, $Q$ - probability distribution of the regression.
 
\begin{align}
    \mathcal{L}(y, y_{\text{pred}}) &= (1 - \alpha) \cdot L1(y, y_{\text{pred}}) + \alpha \cdot D_{\text{KL}}(P \parallel Q) 
    \label{loss_function}
\end{align}

\section{Experiments and results discussion}

We start the experiments with training a recurrent neural network, which has been used as an R-wave peak detector. We use for this the public MIT-BIH dataset, followed by an evaluation of the model's generalizability on the data sets: Fantasia and our obtained form Aidmed ECG. The network achieved the following quality metrics:
On the Fantasia dataset: $acc = 0.999$, $F1 = 0.989$
On our dataset: $acc = 0.998$, $F1 = 0.968$
These results indicate a high generalization capability of the model. 
The detector was then used to increase the number of channels of the training data for the respiratory regression model.

The respiratory regression model was trained for 30 epochs with a batch size parameter of $batch\_size = 32$. The training dataset consisted of signal windows of 60s length, where the respiratory rate ranged within $<6,9>$ breaths per minute (BPM). The Adam algorithm was used with a hyperparameter of $lr = 1e-3$, a \textit{learning scheduler} MultiStepLR with parameters $step\_size = 15$ and $\gamma = 0.1$, and early stopping settings of $patience = 3$, $min\_delta = 1e-5$. However, the training time was not shortened. In the final epoch, the model achieved NCC metric values of $0.938$ for the training data and $0.9436$ for the validation data.

\begin{figure}[!ht]
	\centering
	\includegraphics[scale=0.5]{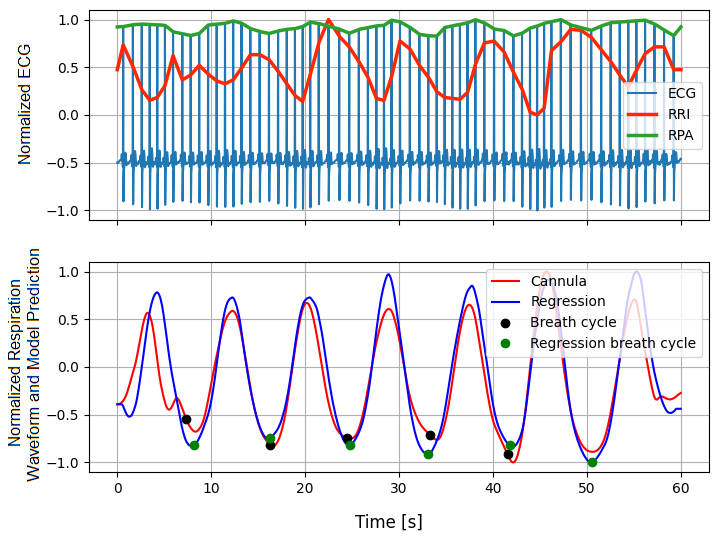}
	\caption{Example of respiratory rate prediction based on ECG signal}
	\label{img/ecg_to_resp_example}
\end{figure}

Fig. \ref{img/ecg_to_resp_example} presents an example from the last training epoch, illustrating the performance of our algorithm. The first signal shows the model's input data, consisting of normalized ECG, RRI, and RPA channels. The second signal visualizes the normalized respiratory and regression values, along with the cycle separation points for these signals.

The tests were conducted using signal windows of 180s in length, with a constant breathing frequency, as well as two samples of 235s in length with mixed breathing frequencies. Three subsets were extracted from the dataset with a constant respiratory rate (RR): 12 examples with low RR ($<4,6>$ BPM), 21 examples with medium RR ($<7,11>$ BPM), and 8 examples with high RR ($<12,\infty>$ BPM). 

Table \ref{table/resp_testset_stats} summarizes the metric values used to evaluate the quality of the designed neural model. The metric "total waveforms duration" describes the total duration of waveforms for each test subset, "average loss function" refers to the loss function for the criterion described in \ref{loss_function}, "average NCC" represents the mean normalized cross-correlation score, and "average RPE" is the mean relative percentage error. The latter is computed between the normalized reference and regression values in the range $<-1,1>$ using the formula
\ref{rpe}.

\begin{equation}
\text{RPE} = \frac{\sum_{i=1}^{N} |y_{\text{pred}}^{(i)} - y_{\text{true}}^{(i)}|}{\sum_{i=1}^{N} |y_{\text{true}}^{(i)}| + \epsilon} \times 100
\label{rpe}
\end{equation}

\renewcommand{\arraystretch}{1.2}

\begin{table}[!t]
\centering
\caption{Test metric values within different ranges of BPM}
\begin{tabularx}{\textwidth}{|X|c|c|c|c|}
\hline
\textbf{Metric} & \textbf{Medium RR} & \textbf{Low RR} & \textbf{High RR} & \textbf{Mixed RR}\\
\hline
Total waveforms duration [min]   & 63  & 36  & 24 & 7,83 \\
\hline
Average breaths per minute [BPM] & 8,30   & 4,19   & 13,46 & 8,55 \\
\hline
Average loss function            & 0,0056  & 0,0098  & 0,0241 & 0,0141\\
\hline
Average NCC                      & 0,95 & 0,89  & 0,47 & 0,84  \\
\hline
Average RPE [\%]                 & 30,20 & 35,52 & 83,00 & 49,80 \\
\hline
\end{tabularx}
\label{table/resp_testset_stats}
\end{table}

Next, we use the \textit{NeuroKit} package to compare the performance of the developed algorithm with a classical approach. The function \textit{nk.ecg\_rsp} was used, which implements the Van Gent algorithm \cite{van2019heartpy}
to estimate respiratory rhythm.

Table \ref{table/resp_comparison} presents a numerical comparison of the quality of the obtained EDR. The best values of the NCC and L1 metrics are highlighted in bold.

\begin{table}
\caption{Comparison of our algorithm (neural) with NeuroKit }
\begin{tabularx}{\textwidth}{|Y|Y|Y|Y|Y|}
\hline 
\textbf{Respiratory Rate [BPM]} & \textbf{Algorithm} & \textbf{Average NCC} & \textbf{Average RPE [\%]} \\
\hline
\multirow{2}{*}{Low <4,6>} & NeuroKit          & 0,31          & 82,86 \\ 
                     & \textbf{Neural}   & \textbf{0,89} & \textbf{35,52} \\
\hline
\multirow{2}{*}{Medium <7,11>} & NeuroKit          & 0,89          & 49,53 \\ 
                     & \textbf{Neural}   & \textbf{0,95} & \textbf{30,20} \\
\hline
\multirow{2}{*}{High <12,$\infty$>} & \textbf{NeuroKit} & \textbf{0,74} & \textbf{69,65} \\ 
                      & Neural            & 0,47          & 83,00 \\
\hline
\multirow{2}{*}{Mixed} & NeuroKit        & 0,63          & 89,92 \\ 
                       & \textbf{Neural} & \textbf{0,84} & \textbf{49,80} \\
                       \hline
                       \hline
\multirow{2}{*}{Average } & NeuroKit        & 0,64          & 72,99\\ 
                       & \textbf{Neural} & \textbf{0,79} & \textbf{49,63} \\
                       \hline
\multirow{2}{*}{Weighted Average} & NeuroKit        & 0,69          & 64,45\\ 
                       & \textbf{Neural} & \textbf{0,84} & \textbf{42,42} \\ 
\hline
\end{tabularx}
\label{table/resp_comparison}
\end{table}

As it can be seen from the results presented in Table \ref{table/resp_comparison}, our neural network architecture performs respiratory waveform regression with significantly higher accuracy compared to the classical algorithm. For 3 out of 4 test subsets, the model achieved better average L1 and NCC metrics. The neural network exhibited lower regression accuracy only for high-frequency waveforms, which is a consequence of the training methodology. Due to information found in publications regarding the RSA phenomenon, training examples were limited, as described in Section \ref{section/method}. At breathing frequencies above around 0.12 Hz (approximately 7.2 breaths per minute), RSA starts vanishing. With very fast breaths, such as during hyperventilation, the heart stops dynamically responds to the respiratory cycle, and RSA may become inconsistent \cite{Niizeki_K}.

\section{Conclusions and future works}
 
The paper describes a neural algorithm for reconstructing respiratory waveforms based on the ECG signal while leveraging the RSA phenomenon. The use of a recurrent network for R-wave peak detection enabled the incorporation of additional information into the single-channel input data, allowing for the training of an effective neural algorithm. The analysis of experimental results indicated the superiority of the proposed method over the classical algorithm in terms of regression accuracy, as confirmed by the obtained metrics.

Due to the decreasing correlation between breathing and heart rhythm at high RR values, the training data range was limited to respiratory frequencies between 6 and 9 breaths per minute. However, testing was conducted on samples representing the full BPM range <4, $\infty$>. As a result, the model's performance deteriorated for extreme values outside the training range. 
In future research, we plan to train the neural model using an unfiltered dataset that includes a broader range of samples to improve its generalization. Experimental results confirmed the model's effectiveness within the medium BPM range. For lower respiratory frequencies, a significant drop in accuracy was observed, which may be caused by increased noise in the RRI channel and the small number of training examples representing this range. A particularly noticeable drop in performance occurred for data with very high RR values, which is associated with a significant vanishing of the RSA phenomenon and reduced variability in the RRI signal.
It is worth noting that the physiological distribution of respiratory frequencies in humans can significantly impact the performance of general models, 
thus adding personalization to the model training may increase its performance.
In addition, we plan to expand the algorithm to use data from wearable devices and include signals with low and high respiratory frequencies. We also aim to define a personal heart rate sensitivity threshold to breathing frequency and analyze the impact of factors like gender and age.
Since this threshold may vary daily, we also plan to identify predictive factors based on the subject’s condition. 
Specifically, we intend to assess ANS branches activity 
using measurements of eyes activity (e.g., saccadic movements, pupil dilation) or EEG signals, which should enable the prediction of these threshold variations.

\section*{Acknowledgments}
The work was founded partially by 
the Gdańsk University of Technology Grant
Americium 3/1/2024/IDUB/II.1b/Am.
The authors would like to thank Aidmed company (\url{https://www.aidmed.ai})
for providing devices for the experiments.

\bibliography{bibliography}
 
\end{document}